# Predicting Station-level Hourly Demand in a Large-scale Bike-sharing Network: A Graph Convolutional Neural Network Approach

Lei Lin [a], Zhengbing He [b] and Srinivas Peeta [c*]

[a] *NEXTRANS Center,*

*Purdue University, West Lafayette, IN 47906, USA*

[b] *College of Metropolitan Transportation,*

*Beijing University of Technology, China*

[c] *School of Civil and Environmental Engineering, and H. Milton Stewart School of Industrial and Systems Engineering*

*Georgia Institute of Technology, Atlanta, GA 30332, USA*

**ABSTRACT**

This study proposes a novel Graph Convolutional Neural Network with Data-driven Graph Filter (GCNN-DDGF) model that can learn hidden heterogeneous pairwise correlations between stations to predict station-level hourly demand in a large-scale bike-sharing network. Two architectures of the GCNN-DDGF model are explored; $GCNN_{reg}$-DDGF is a regular GCNN-DDGF model which contains the convolution and feedforward blocks, and $GCNN_{rec}$-DDGF additionally contains a recurrent block from the Long Short-term Memory neural network architecture to capture temporal dependencies in the bike-sharing demand series. Furthermore, four types of GCNN models are proposed whose adjacency matrices are based on various bike-sharing system data, including Spatial Distance matrix (SD), Demand matrix (DE), Average Trip Duration matrix (ATD), and Demand Correlation matrix (DC). These six types of GCNN models and seven other benchmark models are built and compared on a Citi Bike dataset from New York City which includes 272 stations and over 28 million transactions from 2013 to 2016. Results show that the $GCNN_{rec}$-DDGF performs the best in terms of the Root Mean Square Error, the Mean Absolute Error and the coefficient of determination ($R^2$), followed by the $GCNN_{reg}$-DDGF. They outperform the other models. Through a more detailed graph network analysis based on the learned DDGF, insights are obtained on the "black box" of the GCNN-DDGF model. It is found to capture some information similar to details embedded in the SD, DE and DC matrices. More importantly, it also uncovers hidden heterogeneous pairwise correlations between stations that are not revealed by any of those matrices.

* Corresponding author, Tel.: +1-404-894-2243.

E-mail address: srinivas.peeta@ce.gatech.edu (Srinivas Peeta)

# 1. Introduction

A typical motorized passenger vehicle emits about 4.7 metric tons of carbon dioxide per year (US EPA, 2016). To decrease tailpipe emissions, reduce energy consumption and protect the environment, as of December 2016, roughly 1,000 cities worldwide have started bike-sharing programs (Wikipedia, 2017). Bike sharing can also help to solve the first mile and last mile problems (J.-R. Lin et al., 2013). By providing a connection to other transportation modes, bike usage can seamlessly enable individual trips consisting of multiple transportation modes. Hence, bike sharing is becoming an important component of a modern, sustainable and efficient multi-modal transportation network.

In general, distributed bike-sharing systems (BSSs) can be grouped into two types, dock-based BSS and non-dock BSS. In dock-based BSS, the bikes are rented from and returned to the docking stations. Examples of this BSS type can be found in US cities such as New York City, San Francisco, Chicago and Washington D.C. A non-dock BSS is designed to provide more freedom and flexibility to travelers in terms of access and bike usage. In contrast to dock-based BSS, riders are free to leave bikes wherever they want (Xu et al., 2018). However, this can lead to operational challenges such as blocked sidewalks, etc., in addition to the need to move and restock them at stations. Non-dock BSSs have been deployed in many cities in China by companies such as Ofo and Mobike in 2017, and have rapidly become a popular travel mode. By September 2017, there were 15 bike-sharing programs in operation in Beijing, China that deployed over 2.3 million bikes (Sina, 2017).

While bike sharing can greatly enhance urban mobility as a sustainable transportation mode, it has key limitations due to the effects of fluctuating spatial and temporal demand. As pointed out by many previous studies (Chen et al., 2016; Li et al., 2015; Lin, 2018; Zhou, 2015), it is common for BSSs with fixed stations that some stations are empty with no bikes to check out while others are full precluding bikes from being returned at those locations. For non-dock BSSs, enhanced flexibility poses even more challenges to ensure bike availability at some places and prevent surplus bikes from blocking sidewalks and parking areas. For both types of BSSs, accurate bike-sharing demand predictions can help operators to generate optimal routes and schedules for rebalancing to improve the efficacy of BSSs.

Accurate station-level bike-sharing demand prediction, which is necessary for dynamic bike rebalancing (Caggiani et al., 2018; Liu et al., 2018; Pal and Zhang, 2017), is also very challenging. Most previous studies have built a demand prediction model for each station separately, but failed to utilize hidden correlations between stations to enhance prediction performance (Faghih-Imani et al., 2014; Rixey, 2013; Yang et al., 2016). For example, if a bike station near a subway exit has high demand during the peak period, another one close to it may also have high demand during that period. Furthermore, building a separate demand prediction model for each station is time consuming if the network is large (with hundreds of stations).

Due to its spatio-temporal nature, the station-level bike-sharing demand prediction problem has conceptual similarities with other problems in the transportation domain, e.g., spatio-temporal delay pattern analysis (Z. Zhang et al., 2017; Zhang and Lin, 2017) and short-term traffic prediction in a road network (Min and Wynter, 2011; Vlahogianni et al., 2007). Vlahogianni et al. (2007) utilized temporal volume data from sequential detectors at the same link to improve the prediction accuracy of a multi-layer perceptron (MLP) model. Min and Wynter (2011) built a multivariate spatio-temporal auto-regressive model (MSTAR) to predict network-wide volume and speed considering neighboring links' effect on each link's traffic prediction. The neighboring links of a given link are pre-defined based on whether they are reachable with historical average link speed during a specified period. Unfortunately, these approaches are not applicable to capture the heterogeneous correlations between stations for bike-sharing demand prediction. Two bike-sharing stations that are not on the same link may still have similar demand patterns. Also, a station being reachable from another station in a given period does not indicate that its bike-sharing demand will be high.

More recently, deep learning models have been used to solve transportation problems such as spatio-temporal transportation prediction. Unlike traditional machine learning algorithms, deep learning models are good at representation learning which means they require little effort to manually extract features from raw data (Goodfellow et al., 2016), and their performance has been very promising when enough data is available to train these models. Convolutional Neural Network (CNN), which is a state-of-the-art deep learning model, has been applied for large-scale bike-sharing flow prediction (J. Zhang et al., 2017). CNN is defined and can be applied straight-forwardly for data domains with regular grids such as images. It can also identify correlations among grids with various localized filters or kernels, and these shift-invariant filters are learned from data automatically.

To apply CNN for large-scale spatio-temporal transportation prediction, some data preprocessing work is necessary. Zhang et al. (2017) split the whole city into grids with a pre-defined grid size and calculated the bike-sharing demand for each grid. The demand data, represented using grid maps, was converted into images by defining a color scale. These images were then taken as the input to the CNN model, and the correlations among grids were learned through localized filters. Note that this approach is still not applicable for station-level bike-sharing demand prediction because if the grid size is set too large, multiple stations will be covered by the same grid and fail to satisfy the required granularity. Conversely, if the grid size is as small as one station, the huge image matrix with redundant zero elements will increase the computational burden heavily.

For data lying on irregular or non-Euclidean domains (such as user data in social networks and gene data in biological regulatory networks), the graph has been applied as a main structure to encode the heterogeneous pairwise correlations and complex geometric structures in data (Defferrard et al., 2016). With bike stations as vertices, a bike-sharing network can be represented as a graph. Each station vertex has a feature vector consisting of historical hourly bike-sharing demand values, and an adjacency matrix can be defined to encode the pairwise correlations between stations. A novel deep learning model Graph CNN (GCNN) has been recently proposed to conduct classification and prediction tasks with these kinds of data (Defferrard et al., 2016; Kipf and Welling, 2016). However, once again, the pre-definition of the adjacency matrix is very difficult for the bike-sharing network.

This study proposes a novel deep learning model labeled the Graph Convolutional Neural Network with data-driven graph filter (GCNN-DDGF) model. Our main contributions are summarized as follows: First, it does not require the predefinition of an adjacency matrix, and thus can learn the hidden correlations between stations automatically. Two possible architectures of the GCNN-DDGF model are explored, namely, $GCNN_{reg}$-DDGF and $GCNN_{rec}$-DDGF. The former is a regular GCNN-DDGF model which mainly consists of two types of blocks, the convolution block and the feedforward block. The latter captures temporal dependencies in the bike-sharing demand series by introducing one more block, the recurrent block from the Long Short-term Memory (LSTM) neural network (Hochreiter and Schmidhuber, 1997). To the best of our knowledge, this is the first study to propose a deep learning model to predict station-level hourly demand by utilizing the underlying correlations between stations.

Second, for comparison, we also newly propose four GCNNs, built based on a bike-sharing graph with stations as vertices, where the adjacency matrices are pre-defined using one of the following BSS data: the Spatial Distance matrix (SD), the Demand matrix (DE), the Average Trip Duration matrix (ATD) and the Demand Correlation matrix (DC). The six types of GCNN models as well as seven benchmark models are built on a dataset from the Citi BSS in New York City, which includes 272 stations and over 28 million transactions from 2013 to 2016.

Root Mean Square Error, Mean Absolute Error and coefficient of determination ($R^2$) criteria are calculated. The results show that the performance of GCNN models with pre-defined adjacency matrices is impacted by the adjacency matrix quality, and the novel GCNN$_{rec}$-DDGF performs better than all other models because it captures hidden heterogeneous correlations between stations and the temporal dependencies in the bike-sharing demand series.

Our third main contribution is that we address a well-known issue for deep learning models which is they work like a "black box". In this study, we conduct detailed graph network analysis based on the learned DDGF to understand the "black box" of the GCNN$_{reg}$-DDGF model. Vertex weighted degree analysis shows that a larger weighted degree for a vertex indicates that the DDGF seeks to gather information from more stations to improve the hourly demand prediction for it. Further, the analysis indicates the DDGF not only captures some of the same information that exists in the SD, DE and DC matrices, but also uncovers more underlying correlations between stations than these matrices. This illustrates another key benefit of the GCNN-DDGF model.

To summarize, the contributions of this study include, but are not limited to:

- Proposing a novel GCNN-DDGF model that can automatically learn hidden heterogeneous pairwise correlations between stations to predict station-level hourly demand.

- Two architectures of the GCNN-DDGF model, GCNN$_{reg}$-DDGF and GCNN$_{rec}$-DDGF are explored. Their prediction performances are compared with four GCNN models with pre-defined adjacency matrices and seven benchmark models. The proposed GCNN$_{rec}$-DDGF outperforms all of these models.

- Graph network analysis is conducted on the learned DDGF, which shows the DDGF can capture similar information that is embedded in the SD, DE and DC matrices, and extra hidden heterogeneous pairwise correlations between stations.

The rest of the paper is organized as follows. Section 2 reviews the literature on bike-sharing demand prediction models and GCNN models. The methodology of the GCNN model is presented in detail in Section 3, including the predefinition of the four types of adjacency matrices and the DDGF approach. The Citi BSS dataset is then introduced in Section 4, as are the data preprocessing procedures. Section 5 compares the prediction performance of the various models and explicitly analyzes the learned DDGF. Finally, some concluding comments and future research directions are summarized in Section 6.

## 2. Literature Review

*2.1. Bike-sharing Demand Prediction Models*

A lot of attention has been paid to the bike-sharing demand prediction problem. Based on spatial granularity, there are three groups of prediction models in the literature: city-level, cluster-level, and station-level.

*2.1.1. City-level Bike-sharing Demand Prediction*

For the city-level group, the objective is to predict the bike usage for a whole city. In 2014, Kaggle, the world's largest platform for predictive modeling and analytics competitions, invited participants to forecast the total hourly demand in the Capital Bike Share program in Washington, D.C.(Kaggle, 2015). Giot and Cherrier (2014) made demand predictions for the next 24 hours with a city-level granularity for the Capital Bike Share system. They tested multiple machine learning algorithms such as Ridge Regression, Adaboost Regression, Support Vector Regression, Random Forecast Tree and Gradient Boosting Regression Tree and showed that the first two outperformed the others. Although predicting the total city-level rentals from all bike-sharing stations simplifies the problem greatly, it does not contribute to solving the bike re-balancing problem among stations. Further, more detailed transaction data collected by BSSs, such as trip duration, origin, destination, check in/out time, user information and so on, are not fully utilized in the city-level models.

*2.1.2. Cluster-level Bike-sharing Demand Prediction*

The assumption behind the cluster-level group is that some correlations exist among stations based on their geographical locations and temporal demand, and that the total demand of these stations can be predicted as a cluster. For example, bike usage patterns are usually similar for a small cluster of stations near a residential area during the morning rush hours. If the cluster-level predictions are accurate enough, one can always find an available bike within that cluster. A few studies have tried to identify these kinds of spatio-temporal clusters among stations. Zhou (2015) applied the Community Detection algorithm and the Agglomerative Hierarchical Clustering method to group similar bike flows and stations in the Chicago BSS. The study verified that the bike usage patterns of the clusters are different by day, user, directional and land use profiles. Bao et al. (2017) applied the K-means clustering algorithm and the Latent Dirichlet Allocation (LDA) to discover the hidden bike-sharing travel patterns and trip purposes. A few other studies have also applied clustering algorithms to predict cluster-level bike-sharing demand. Li et al. (2015) proposed a bike-sharing demand prediction framework that introduces a Bipartite Station Clustering algorithm to group individual stations. The whole city bike-sharing demand is predicted based on the Gradient Boosting Regression Tree and later split across clusters based on a Multi-similarity-based Inference model. Chen et al. (2016) pointed out that the clustering of stations should be updated based on temporal and weather factors, and social and traffic events. They proposed a Geographically-constrained station Clustering method over a weighted correlation network to dynamically group stations with similar bike usage patterns. Then, they estimated the cluster-level rental and return numbers with the cluster's average value adjusted by an inflation rate.

*2.1.3. Station-level Bike-sharing Demand Prediction*

Station-level bike-sharing demand prediction is more challenging (Chen et al., 2016; Li et al., 2015), and has attracted considerable interest recently. Among previous studies on station-level bike-sharing demand prediction, few have considered underlying spatial or temporal correlations between stations to improve the prediction performance. Rixey (2013) built linear regression models for predicting monthly rentals by station in three BSSs: Capital Bike Share, Denver B-Cycle, and Nice Ride MN systems. Independent variables such as demographic factors and built environment factors are extracted based on a 400-meter buffer around each station (Rixey, 2013). Faghih-Imani et al. (2014) similarly built linear mixed models to predict the hourly bike-sharing demand by station based on a two-day dataset for the BIXI bicycle sharing system in Montreal. Similarly, a 250-meter buffer is set up for each station to generate explanatory variables in the models. Yang et al. (2016) proposed a probabilistic mobility model which considers the previous check-out records and trip durations to estimate the future check-in numbers at each station. However, for bicycle check-out or demand predictions, they applied the Random Forest tree algorithm for each station without leveraging spatial or temporal correlations between stations.

*2.2. Graph Convolutional Neural Network*

Considering CNN is mainly designed to extract highly meaningful statistical patterns and learn local stationary structures presented in data such as image and video, some emerging studies have proposed the graph CNN (GCNN) for data lying on irregular domains. One approach for filtering in GCNN models is to utilize signal processing theory in graphs (Bruna et al., 2013; Sandryhaila and Moura, 2013; Shuman et al., 2013). Given an undirected and connected graph, each vertex has a signal or feature vector, and an adjacency matrix (weighted or binary) is defined where each entry encodes the degree of relation between signal vectors at two vertices (Sandryhaila and Moura, 2013). The Laplacian matrix (Shuman et al., 2013) or the adjacency matrix (Sandryhaila and Moura, 2013) of the graph can be decomposed to form the Fourier basis. The Graph Fourier Transform is then performed to convert the signal data from vertex domain to frequency domain. Then, graph spectral filtering can be conducted to amplify or attenuate the contributions of some of the components (Shuman et al., 2013). Defferrard et al. (2016) proposed a fast localized spectral filtering approach and applied their GCNN model for text classification with promising results. Kipf and Welling (2016) applied a similar localized spectral filter, which is further simplified using the first-order approximation. Multiple citation datasets were used to test their GCNN for semi-supervised learning, and it is shown to outperform other such models in both efficiency and accuracy. However, as pointed out by the authors, critical shortcomings of the GCNN are the need to create the graph artificially and predefine the adjacency matrix. And the quality of the input graph is of paramount importance (Defferrard et al., 2016; Kipf, 2016).

## 3. Methodology

This section first describes the GCNN and the spectral filtering methodology. After that, multiple approaches to define the adjacency matrix in a bike-sharing network are discussed. Our data-driven approach to learn the graph spectral filter is then introduced.

### 3.1. Graph Convolutional Neural Network

Suppose we have a graph $G = (V, x, \mathcal{E}, A)$, where $V$ is a finite set of vertices with size $N$, signal $x \in \mathbb{R}^N$ is a scalar for every vertex, $\mathcal{E}$ is a set of edges, $A \in \mathbb{R}^{N \times N}$ is the adjacency matrix, and entry $A_{ij}$ encodes the connection degree between the signals at two vertices. A normalized graph Laplacian matrix is defined as

$$L = I_N - D^{-1/2} A D^{-1/2} \tag{1}$$

where $I_N$ is the identity matrix, and $D \in \mathbb{R}^{N \times N}$ is a diagonal degree matrix with $D_{ii} = \sum_j A_{ij}$.

$L$ is a real symmetric positive semidefinite matrix which can be diagonalized as

$$L = U \Lambda U^T \tag{2}$$

where $U = [u_0, u_1, ..., u_{N-1}]$; $\Lambda = diag([\lambda_0, \lambda_1, ..., \lambda_{N-1}])$; $\lambda_0, \lambda_1, ..., \lambda_{N-1}$ are the eigenvalues of $L$, and $u_0, u_1, ..., u_{N-1}$ are the corresponding set of orthonormal eigenvectors.

*3.1.1. Spectral Convolution on Graph*

A spectral convolution on the graph is defined as follows:

$$g_\theta * x = U g_\theta(\Lambda) U^T x \qquad (3)$$

where $g_\theta(\Lambda)$ is a function of the eigenvalues of $L$.

A form of polynomial filters has been used in a few studies (Defferrard et al., 2016; Kipf and Welling, 2016; Shuman et al., 2013):

$$g_\theta(\Lambda) = \sum_{k=0}^{K} \theta_k \Lambda^k \qquad (4)$$

With the polynomial filters, Equation (3) can be written as:

$$g_\theta * x = \sum_{k=0}^{K} \theta_k U \Lambda^k U^T x = \sum_{k=0}^{K} \theta_k L^k x \qquad (5)$$

Hammond et al. (2011) shows that the entry $(L^k)_{i,j} = 0$ when the shortest path distance between vertices $i$ and $j$ is greater than $k$. Therefore, this type of filter is also known as the $K$-localized filter. The physical meaning of the graph spectral convolution is that it combines the signal at the central vertex with the signals at vertices that are a maximum of $K$ steps away.

To improve the computational efficiency, Kipf and Welling (2016) simplified the calculation of $g_\theta * x$ by using only the first-order polynomial:

$$g_\theta * x \approx \widetilde{D}^{-\frac{1}{2}} \widetilde{A} \widetilde{D}^{-\frac{1}{2}} x \theta \qquad (6)$$

where $\theta \in \mathbb{R}$; $\widetilde{A} = A + I_N$ is the summation of the adjacency matrix of the undirected graph $A$ and the identity matrix $I_N$. That is, $\widetilde{A}$ is the adjacency matrix of an undirected graph where each vertex connects with itself; $\widetilde{D}_{ii} = \sum_j \widetilde{A}_{ij}$.

Generalizing this convolution calculation to a signal $X \in \mathbb{R}^{N \times C}$ where each vertex $v_i$ has a $C$-dimensional feature vector $X_i$,

$$Z = \widetilde{D}^{-\frac{1}{2}} \widetilde{A} \widetilde{D}^{-\frac{1}{2}} X \Theta \qquad (7)$$

where $\Theta \in \mathbb{R}^{C \times F}$ is a matrix of filter parameters, and $Z \in \mathbb{R}^{N \times F}$ is the convolved signal matrix.

*3.1.2. Layer-wise Calculation*

Suppose the GCNN model has layers $0, 1, ..., m$ from the input to the output. Each vertex of the graph at each layer $l$, $l = 0, 1, ..., m$, has a feature vector of length $C^l$. For each layer $l$, $l = 1, ..., m-1$, the GCNN model $f(X, A)$ propagates from the input to the output with the rule:

$$H^l = \sigma(Z^{l-1} W_T^l) = \sigma\left(\widetilde{D}^{-\frac{1}{2}} \widetilde{A} \widetilde{D}^{-\frac{1}{2}} H^{l-1} \Theta^{l-1} W_T^l\right) \qquad (8)$$

where $Z^{l-1}$ is the convolved signal matrix in the $(l-1)^{th}$ layer, $W_T^l \in \mathbb{R}^{F^{l-1} \times C^l}$ is a layer-specific trainable weight matrix, $\sigma(\cdot)$ is an activation function (in this study, we apply the ReLU activation function), and $H^l \in \mathbb{R}^{N \times C^l}$ is the matrix of activations in the $l^{th}$ layer, $H^0 = X$.

The product of $\Theta^{l-1} \in \mathbb{R}^{C^{l-1} \times F^{l-1}}$ and $W_T^l \in \mathbb{R}^{F^{l-1} \times C^l}$ can be learned by the neural network as one matrix $W^l \in \mathbb{R}^{C^{l-1} \times C^l}$; therefore, (8) can be simplified as:

$$H^l = \sigma\left(\widetilde{D}^{-\frac{1}{2}} \widetilde{A} \widetilde{D}^{-\frac{1}{2}} H^{l-1} W^l\right) \tag{9}$$

For the output layer $m$, the result is:

$$H^m = \widetilde{D}^{-\frac{1}{2}} \widetilde{A} \widetilde{D}^{-\frac{1}{2}} H^{m-1} W^m \tag{10}$$

where $W^m \in \mathbb{R}^{C^{m-1} \times C^m}$ are the weight parameters to be learned, and $H^m \in \mathbb{R}^{N \times C^m}$ are the predictions, e.g., the bike-sharing demand of the $N$ stations for the next hour when $C^m = 1$.

*3.2. GCNN with Pre-defined Adjacency Matrix*

The GCNN model relies on the structure of the graph. The adjacency matrix $\widetilde{A}$ needs to be defined first, with which the graph spectral filter can be approximated. This section proposes four typical data matrices in a BSS to quantify the correlations between stations. Correspondingly, four types of adjacency matrices $\widetilde{A}$ can be constructed.

*3.2.1. Spatial Distance Matrix*

One way to encode the connection between stations is simply through the spatial distance (Shuman et al., 2013). The spatial distance (SD) matrix can be built using the spherical distances with the known latitudes and longitudes of the stations (here the spherical distance between two points is the shortest distance over the Earth's surface). If two stations are spatially close to each other, they are connected in the bike-sharing graph network, and the element of $\widetilde{A}$ is then defined as:

$$\widetilde{A}_{ij} = \begin{cases} 1 & \text{if } DIST_{ij} \leq \kappa_{SD} \\ 0 & \text{if } DIST_{ij} > \kappa_{SD} \end{cases} \tag{11}$$

where $DIST_{ij}$ is the spatial distance between station $i$ and $j$, and $\kappa_{SD}$ is a pre-defined SD threshold.

*3.2.2. Demand Matrix*

This approach makes use of the check in and check out station information in the bike-sharing transaction records. The symmetric demand matrix (DE), which considers the total demand

between stations $i$ and $j$, is built as follows:

$$DE_{ij} = \begin{cases} OD_{ij} + OD_{ji} & if\ i \neq j \\ OD_{ij} & otherwise \end{cases} \quad (12)$$

where $OD_{ij}$ is the aggregated demand from station $i$ to station $j$.

For the bike-sharing graph network, if the total demand between two stations $DE_{ij}$ is higher than a pre-defined threshold $\kappa_{DE}$ the two stations are connected. In this way, a binary $\tilde{A}$ can be built such that $\tilde{A}_{ij} = 1$ if $DE_{ij} \geq \kappa_{DE}$; otherwise $\tilde{A}_{ij} = 0$.

### 3.2.3. Average Trip Duration Matrix

This approach utilizes the trip duration information in the bike-sharing transaction records on the basis of the DE matrix. Each entry $ATD_{ij}$ in an Average Trip Duration matrix (ATD) is defined as:

$$ATD_{ij} = TTD_{ij}/DE_{ij} \quad (13)$$

where $TTD_{ij}$ is the total trip duration of all trips between stations $i$ and $j$; $DE_{ij}$ is an entry in the DE matrix.

Similarly, a binary adjacency matrix is formed such that $\tilde{A}_{ij} = 1$ if $ATD_{ij} \leq \kappa_{ATD}$, otherwise $\tilde{A}_{ij} = 0$. $\kappa_{ATD}$ is a pre-defined threshold. Thereby, the bike-sharing graph network based on the ATD matrix seeks to connect two stations if the average trip duration between them is short.

### 3.2.4. Demand Correlation Matrix

This approach seeks to capture the temporal demand correlations between stations by employing the check-out times of the bike-sharing transaction records. As will be shown in Section 3, based on the check-out times, the bike-sharing transactions are aggregated by hour. Finally, each station has a time series of 26,304 hourly bike-sharing demand values for the three-year period (2013-2016). The Demand Correlation matrix (DC) is defined by calculating the Pearson Correlation Coefficient (PCC) based on the hourly bike demand series between station $i$ and station $j$.

$$DC_{ij} = PCC(h_i, h_j) \quad (14)$$

where $h_i$ and $h_j$ are the hourly bike demand series for stations $i$ and $j$.

In this approach, the bike-sharing graph network is built by connecting two stations with high PCC. Each entry of $\tilde{A}$ is set to 1 if $DC_{ij} \geq \kappa_{DC}$, otherwise $\tilde{A}_{ij} = 0$; where $\kappa_{DC}$ is the DC threshold.

*3.3. GCNN with Data-Driven Graph Filter*

*3.3.1. Data-driven Graph Filter*

The predefinition of the adjacency matrix $\tilde{A}$ is not trivial. The hidden correlations between stations may be heterogeneous. Hence, it may be hard to encode them using just one kind of metric such as the SD, DE, ATD or DC matrix. Now, suppose the adjacency matrix $\tilde{A}$ is unknown; let $\hat{A} = \tilde{D}^{-\frac{1}{2}} \tilde{A} \tilde{D}^{-\frac{1}{2}}$, then (9) becomes:

$$H^l = \sigma(\hat{A} H^{l-1} W^l) \tag{15}$$

where $\hat{A}$ is called the Data-driven Graph Filter (DDGF) which is a symmetric matrix consisting of trainable filter parameters, $\hat{A} \in \mathbb{R}^{N \times N}$, $H^l \in \mathbb{R}^{N \times C^l}$.

Here, the graph filter $\hat{A}$ can be learned from data after training the deep learning model. This DDGF can learn hidden heterogeneous pairwise correlations between stations to improve prediction performance. Such a GCNN model is labeled the GCNN-DDGF. From another perspective, we can view the data-driven graph filtering as filtering in the vertex domain, which avoids the three operations: graph Fourier transform, filtering and inverse graph Fourier transform.

*3.3.2. Architecture Design*

We explore two possible architectures of the GCNN-DDGF. The first, GCNN$_{reg}$-DDGF, contains two types of blocks, the convolution block and the feedforward block. Fig. 1(a) shows an example of it with three stations $i, j$ and $k$. In Layer $(l-1)$, the signal vectors for the three stations are $H_i^{l-1} \in \mathbb{R}^{C^{l-1}}$, $H_j^{l-1}$ and $H_k^{l-1}$. From Layer $(l-1)$ to Layer $l$, two steps are implemented. In step 1, through the convolution block, the signal vector at each station vertex is amplified or attenuated, and linearly combined with signals at other vertices weighted proportionally to the learned degree of their correlations. The signal vectors become $(\hat{A} H^{l-1})_i$, $(\hat{A} H^{l-1})_j$ and $(\hat{A} H^{l-1})_k$. This is shown in Fig. 1(a) through the combination of three different background patterns at each vertex. In step 2, the signal vectors at the vertices of the next layer $l$ are calculated using the traditional feedforward block (the basic block in neural network models) to form the new signal vectors at Layer $l$ $H_i^l \in \mathbb{R}^{C^l}$, $H_j^l$ and $H_k^l$. The dimension of the vector at each vertex changes from $C^{l-1}$ to $C^l$. Suppose the number of layers of GCNN$_{reg}$-DDGF model is $(m+1)$, with layers $0, 1, ..., m$ from the input to the output; then, steps 1 and 2 perform the layer-wise calculation from layer $(l-1)$ to $l$, $l = 1, ..., m$. Note that Fig. 1(a) is a toy example to conceptually illustrate the implementation of the GCNN<sub>reg</sub>-DDGF on three stations. In the New York City application, the model is applied to the whole bike-sharing network.

The second architecture, GCNN<sub>rec</sub>-DDGF, shown in Fig. 1(b), imports an additional block beyond the convolution and feedforward blocks, labeled the recurrent block, from the Long Short-term Memory (LSTM) neural network architecture. The LSTM model is well-suited to capture temporal dependencies in time series data (Hochreiter and Schmidhuber, 1997). It has been successfully applied for natural language text compression, speech/handwriting recognition, etc. In the transportation domain, it was applied for traffic speed prediction (Ma et al., 2015), vehicle trajectory prediction (Lin et al., 2018b) and so on. More recently, the integration of the LSTM architecture with the CNN architecture has been reported to improve large-scale taxi demand predictions by modeling both spatial and temporal relationships (Ke et al., 2017; Yao et al., 2018). Hence, we expect the introduction of the recurrent block in GCNN<sub>rec</sub>-DDGF to potentially improve bike-sharing demand prediction performance.

As shown in Fig. 1(b), at time step $e$, $e = t - T + 1, \dots, t$, we have a signal $x^e \in \mathbb{R}^N$, where $N$ is the number of bike sharing stations (3 in this example). For each $x^e$, the convolution step is conducted through the multiplication $\hat{A}x^e$. These $\hat{A}x^e$, $e = t - T + 1, \dots, t$ are fed into a recurrent block, and the total LSTM cell number in this block is $T$. Each LSTM cell maps the input vector $\hat{A}x^e$ to a hidden state $h^e \in \mathbb{R}^d$, where $d$ is the number of hidden units in each cell. The computation process of the LSTM cell follows standard steps involving the input gate, forget gate, output gate and memory cell state; details can be found in previous studies (Ke et al., 2017; Ma et al., 2015). Finally, the hidden state $h^t$ from the last LSTM cell is taken as the input to a fully-connected feedforward block, the output of which is the bike-sharing demand predictions of all stations at time step $(t + 1)$. There are two hyper-parameters to determine in this architecture: the number of time steps $T$ and the number of hidden units $d$ in each cell.

## 4. Citi Bike-Sharing Demand Dataset

The study dataset includes over 28 million bike-sharing transactions between 07/01/2013 and 06/30/2016, which are downloaded from Citi BSS in New York City (Citi Bike NYC, 2017). Each transaction record includes details such as trip duration, bike check out/in time, start and end station names, start and end station latitudes/longitudes, user ID, user type (Customer or Subscriber), and so on. A few data preprocessing operations are described next.

First, we observe that new stations were being set up from 2013 to 2016. Therefore, only stations existing in all three years are included in the study. Second, some stations were rarely utilized during these three years. Hence, stations with total three-year demand of less than 26,304 (less than one bike per hour) are also excluded. Based on this, 272 stations are considered in the study. The SD matrix is built with the known latitudes and longitudes of these stations. Fig. 2(a) shows these 272 stations on the New York City map. Most of the stations are in southern Manhattan, and only some are located in Brooklyn across the Brooklyn Bridge, Manhattan Bridge, and Williamsburg Bridge.

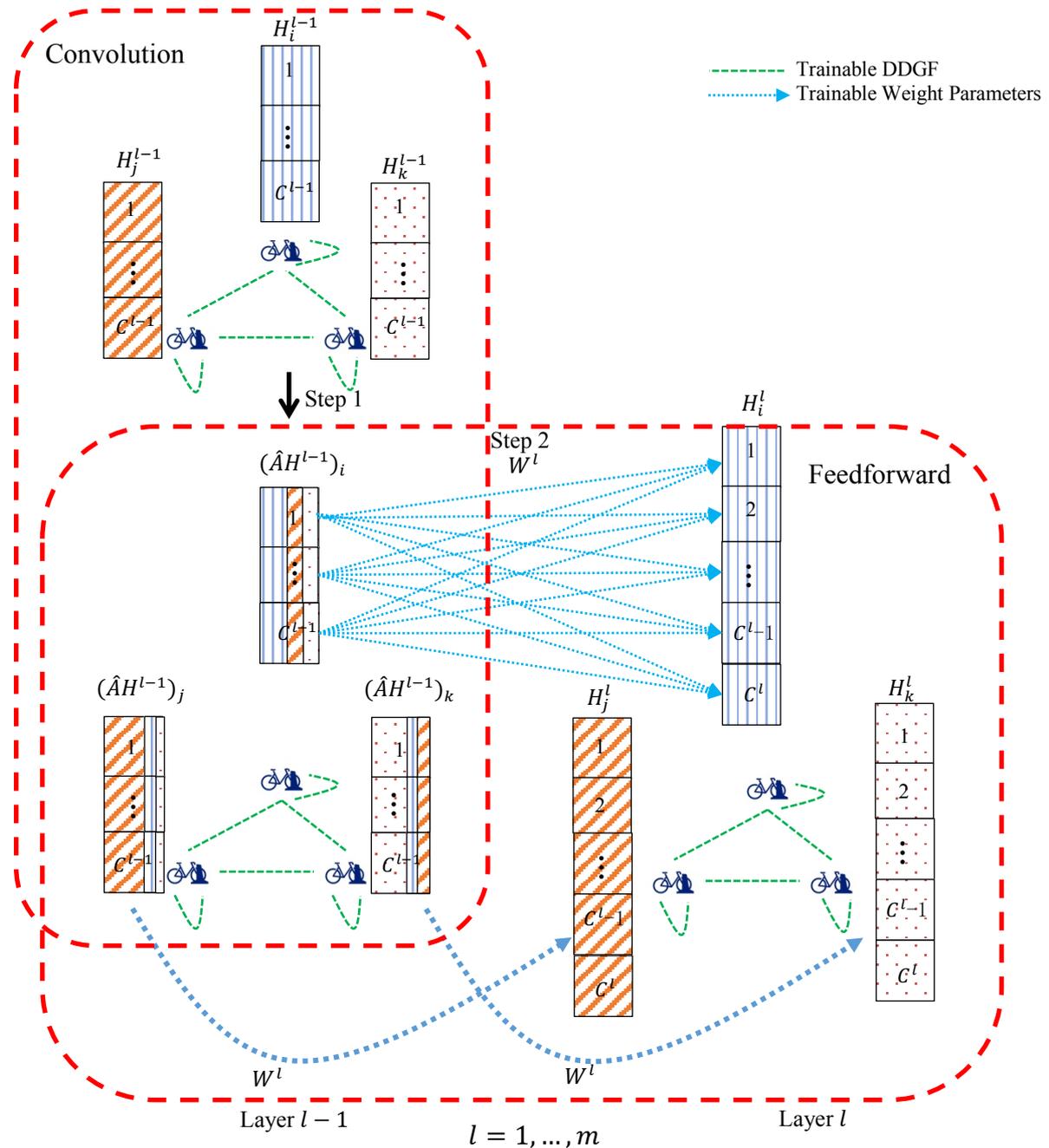

(a)

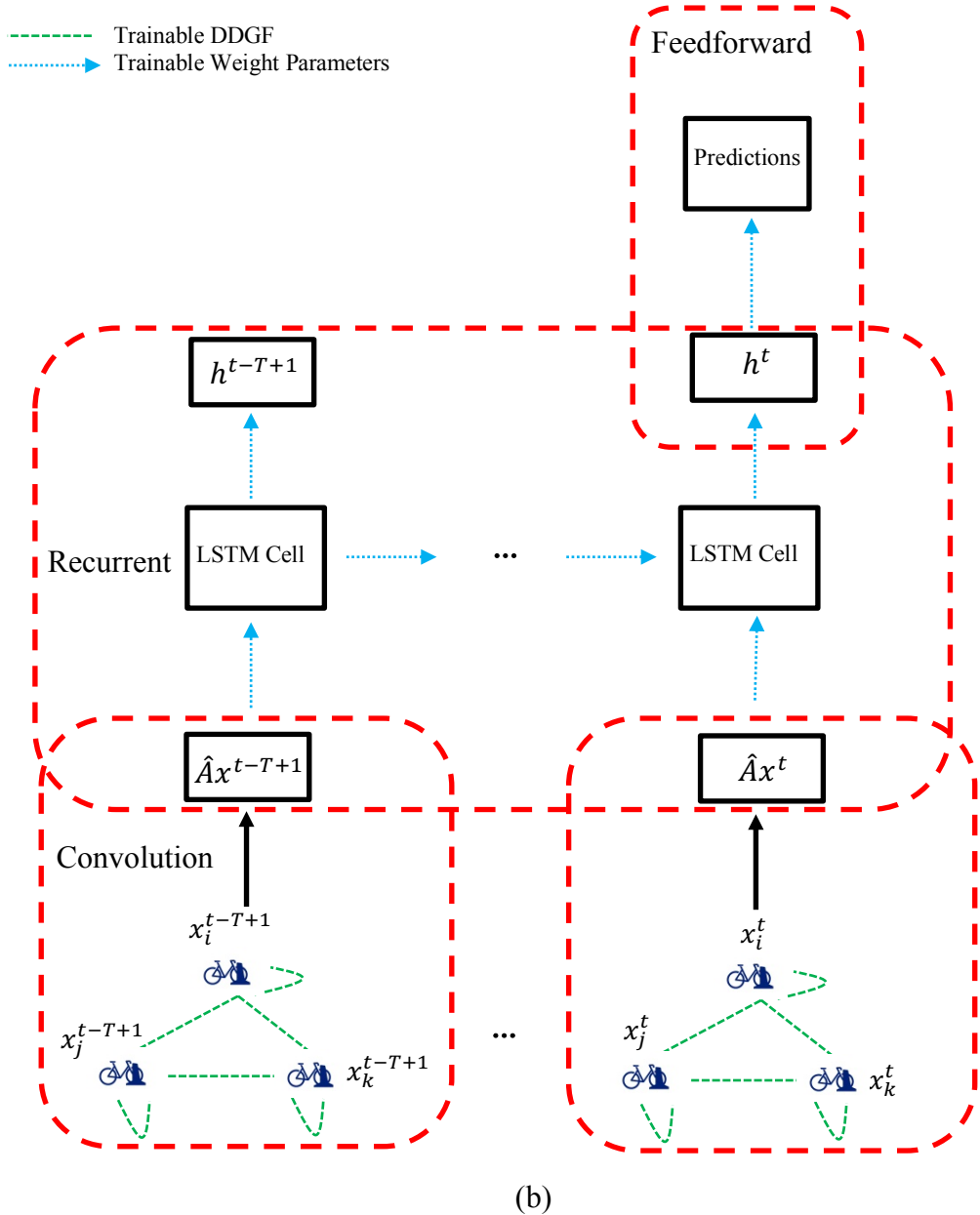

(b)

**Fig. 1.** (a) Illustration of the GCNN$_{reg}$-DDGF architecture, and (b) Illustration of the GCNN$_{rec}$-DDGF architecture.

For each station, 26,304 hourly bike demands are aggregated based on the bike check out time and start station. Fig. 2(b) shows the average hourly bike demand for each of the 272 stations sorted in ascending order, and Fig. 2(c) shows the corresponding standard deviations. It can be noted that some stations are busy, and have an average demand of more than ten bikes per hour. Other stations are less utilized. Further, it can be noted that, in general, a higher average bike demand for a station also implies a larger standard deviation.

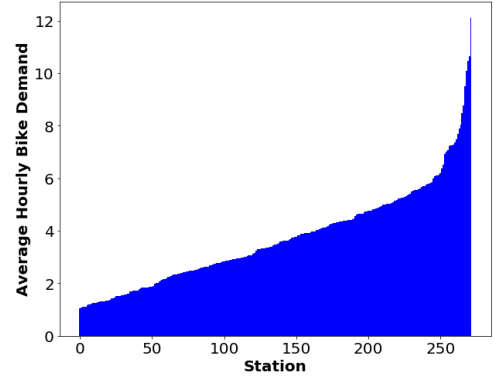

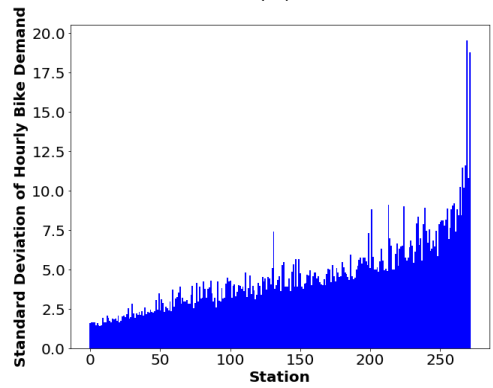

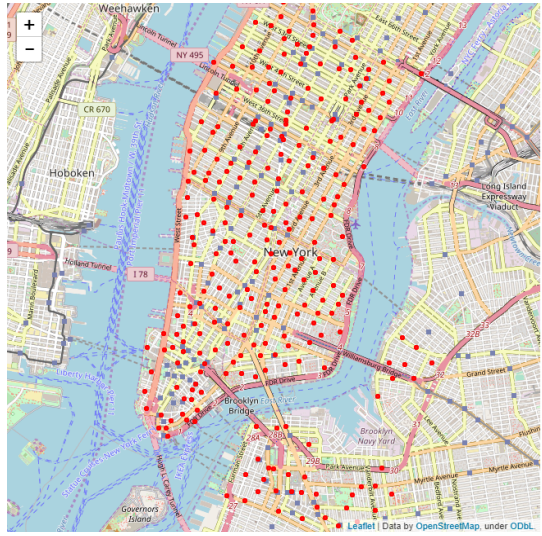

**Fig. 2.** (a) Locations of 272 Bike-Sharing Stations in New York City, (b) Average Hourly Bike Demand by Station, and (c) Standard Deviation of Hourly Bike Demand for the 272 Stations.

For all stations, the first 22,304 hourly bike-sharing demand values are used to train the models, the next 2,000 hourly bike-sharing demand values are included in the validation dataset, and the remaining 2,000 are taken as the testing dataset. The DE, ATD and DC matrices are built only from the training dataset. These matrices are analyzed hereafter.

Fig. 3(a) shows the aggregated demand between stations by distance based on the DE and SD matrices. The highest demand is over 4.5 million trips, when the distance between stations is 1 to 2 miles. The demand drops when the distance is closer (0 to 1 mile), as well as when the distance increases beyond 1-2 miles. The average trip durations by distance based on the ATD and SD are shown in Fig. 3(b). The average duration is about 10 minutes for the trips within 1 mile. It increases with the distance and can take more than 45 minutes when the trips are longer than 5 miles. Note that the actual trip distances are unknown, and here we consider the spatial distances based on the latitudes and longitudes of stations.

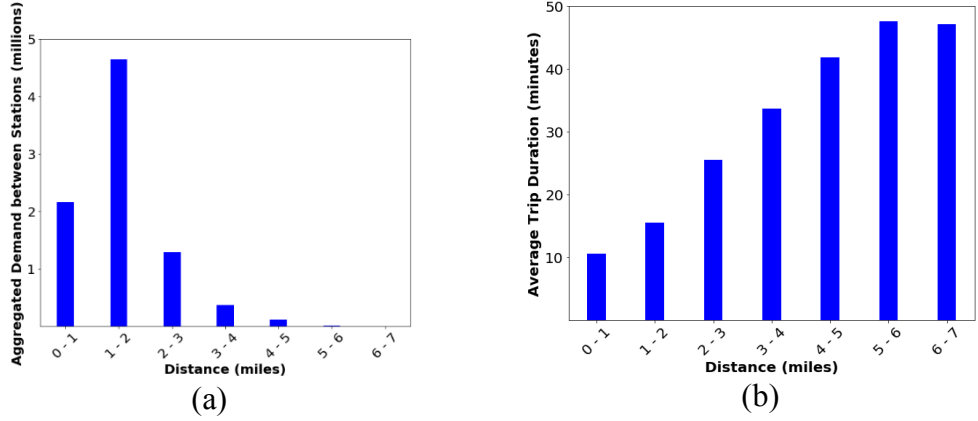

(a)                                                 (b)

**Fig. 3.** (a) Aggregated Demand between Stations by Distance, and (b) Average Trip Duration by Distance.

The 272 × 272 entries in the DC matrix, which are the Pearson correlation coefficients between stations, are shown in Fig. 4(a) The temporal bike demands of the stations are highly correlated. Fig. 4(b) shows the normalized histogram of the demand correlation coefficients. There are no negative coefficients, which implies that the bike-sharing demands of any two stations always move in the same direction, increasing or decreasing together. Also, the histogram is mostly shaped like a Gaussian distribution with mean equal to 0.50. Additionally, 0.5% of the correlation coefficients are very close to 1.

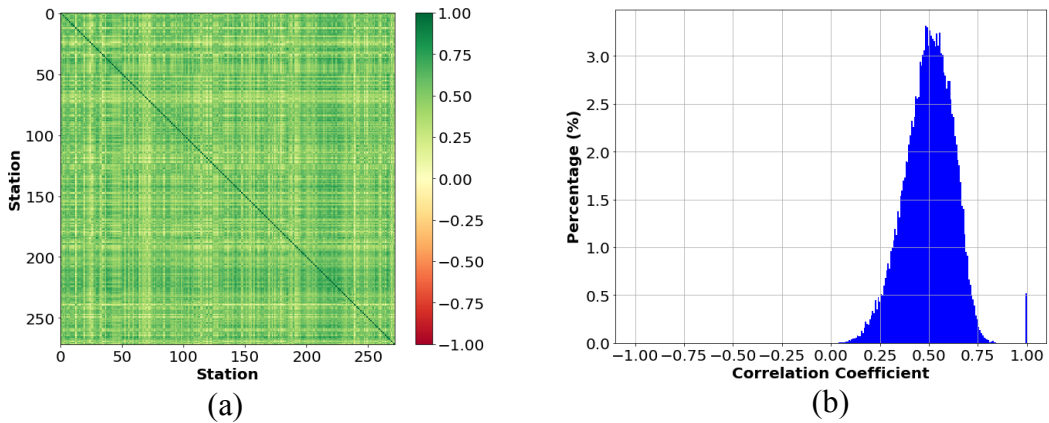

(a)                                                 (b)

**Fig. 4.** (a) Visualization of the DC Matrix, and (b) Normalized Histogram of Demand Correlation Coefficients.

### 5. Model Development and Results

Suppose the bike-sharing demands for all stations in hour $i$ are $x_i \in \mathbb{R}^N$, $N = 272$. Then, using the demand from the previous ($C^0 - 1$) hours, we can construct a feature matrix $X_i \in$

$\mathbb{R}^{N \times C^0}$, $X_i = [x_{i-C^0+1}, \ldots, x_i]$, and the corresponding target vector $y_{i+1} \in \mathbb{R}^N$ which is the vector of bike-sharing demand of all stations in the next hour. The original training dataset is transferred into records $\{(X_i, y_{i+1})\}$, $i = C^0, \ldots, 22,303$. The Min Max normalization is applied to scale the data between 0 and 1. Note that in this study we do not consider the periodicity in historical bike-sharing demand series. Some previous studies related to short-term demand forecasting show that utilizing information on demand from some recent time steps is sufficient to build models that predict well (Ke et al., 2017; L. Lin et al., 2013b; Vlahogianni et al., 2007). The selection of $C^0$ will be described later.

We build six types of GCNN models, GCNN-SD, GCNN-DE, GCNN-ATD, GCNN-DC, GCNN$_{\text{reg}}$-DDGF, and GCNN$_{\text{rec}}$-DDGF, which are labeled based on how the adjacency matrix is generated. Their performance is evaluated using the Root Mean Square Error (RMSE) as the main criterion (J. Zhang et al., 2017):

$$RMSE = \sqrt{\frac{1}{M*N} \sum_i^M \sum_j^N (y_{ij} - P_{ij})^2} \qquad (16)$$

where $M$ is the number of hours, $N$ is the number of stations, and $P_{ij}$ and $y_{ij}$ are the predicted and recorded bike demand in hour $i$ for station $j$, respectively.

*5.1. Hyper-parameter Selection for GCNN models*

In machine learning, hyper-parameters are parameters whose values are set prior to the commencement of the learning process. The traditional way of performing hyper-parameter optimization is grid search, which is to manually specify the subset of hyper-parameter space and perform the search exhaustively. The trained model is evaluated on the validation dataset to determine the optimal hyper-parameters.

Table 1 summarizes the hyper-parameters for the GCNN models. For each parameter, we use $\{start: step: end\}$ to define the search space, e.g., $\{1: 2: 5\}$ represents the space $\{1, 3, 5\}$. The first group of hyper-parameters are used to determine the architecture of the GCNN models. They include the threshold $th$ to form the adjacency matrices from the SD, DE, ATD or DC matrices, and the feature vector length in the input layer $C^0$ and those in the hidden layers 1 and 2: $C^1$ and $C^2$, respectively. Note that we only consider the GCNNs with at most two hidden layers in this study; if $C^2$ is 0, it means the optimal model only needs one hidden layer. For the GCNN$_{\text{rec}}$-DDGF, we have the number of time steps $T$ in the recurrent block and the number of hidden units $d$ in each cell. The second group includes two parameters, learning rate α and mini-batch size $B$, in the classic Stochastic Gradient Descent Algorithm (SGD). The third group is to prevent overfitting. Overfitting means the model is fitting random errors or noises instead of the underlying relationship. With early stopping mechanism, the training algorithm terminates when the performance on the validation dataset has not improved for a pre-specified threshold of iterations $s$. The stored best model on the validation dataset is returned as the final model.

Based on Table 1, the experiments are conducted using Python 3.0 in a Ubuntu 16.04 Linux System with 64 GB RAM and GTX 1080 Graphics Card. The models are developed using TensorFlow, an open-source deep learning neural network library maintained by Google. TensorFlow supports the GTX 1080 GPU which provides strong computation power. The optimal hyper-parameters are listed in Table 2.

**Table 1**

Hyper-parameter Selection in GCNN Models

| Hyper-parameters | | GCNN-SD | GCNN-DE | GCNN-ATD | GCNN-DC | GCNN$_{reg}$-DDGF |
|---|---|---|---|---|---|---|
| Model Architecture | $th$ | {1:2:5} | {500:200:1100} | {10:10:30} | {0.5:0.2:0.9} | N/A |
| | $C^0$ | | | {6:6:48} | | |
| | $C^1$ | | | {20:20:60} | | |
| | $C^2$ | | | {0:20:60} | | |
| | | | | GCNN$_{rec}$-DDGF | | |
| | $T$ | | | {12:12:48} | | |
| | $d$ | | | {25:25:200} | | |
| SGD | $\alpha$ | | | {0.001:0.001:0.015} | | |
| | $B$ | | | {100:100:200} | | |
| Overfitting | $s$ | | | {10:10:100} | | |

**Table 2**

Optimal Hyper-parameters and Performance on the Validation Dataset

| | | GCNN-SD | GCNN-DE | GCNN-ATD | GCNN-DC | GCNN$_{reg}$-DDGF | GCNN$_{rec}$-DDGF |
|---|---|---|---|---|---|---|---|
| Optimal Hyper-parameters | $th$ | 3 | 900 | 20 | 0.9 | N/A | N/A |
| | $C^0$ | 36 | 24 | 36 | 24 | 24 | N/A |
| | $C^1$ | 40 | 40 | 40 | 40 | 40 | N/A |
| | $C^2$ | 0 | 0 | 0 | 0 | 0 | N/A |
| | $T$ | N/A | N/A | N/A | N/A | N/A | 24 |
| | $d$ | N/A | N/A | N/A | N/A | N/A | 100 |
| | $\alpha$ | 0.01 | 0.01 | 0.01 | 0.005 | 0.005 | 0.001 |
| | $B$ | 100 | 100 | 100 | 100 | 100 | 100 |
| | $s$ | 20 | 20 | 20 | 20 | 20 | 100 |
| RMSE on Validation Dataset | | 3.39 | 3.36 | 3.97 | 3.14 | 2.97 | 2.46 |

For the performance on the validation dataset, GCNN-DDGF models learn the optimal graph filter $\hat{A}$ directly from the data, and the RMSEs on validation dataset are 2.97 and 2.46 separately. By contrast, the RMSE of GCNN-DC is 3.14 when only the stations with highly correlated hourly demand series are connected in the graph (the threshold $\kappa_{DC}$ is set as high as 0.9). The GCNN-SD and GCNN-DE have validation RMSEs of 3.39 and 3.36, respectively, and GCNN-ATD performs the worst with a validation RMSE of 3.97. It should be noted here that we also construct graphs with weighted adjacency matrices, whereby the entries in the SD, DE, ATD or DC are directly set as the edge weights in the graphs if they satisfy the threshold requirements. However, the performance of the GCNNs based on these graphs is worse than the GCNNs on graphs with binary adjacency matrices. This verifies insights from previous studies that the adjacency matrix is of paramount importance (Defferrard et al., 2016; Kipf, 2016) for GCNN models, and pre-defining the correlations between stations in a reasonable way is not trivial.

*5.2. Model Comparison on Testing Dataset*

The seven benchmark models for comparison with GCNN models are briefly introduced here as follows:

(1) HA: The Historical Average model predicts the bike sharing demand of the next hour for a station based on the average of historical demands for the same hour at this station in the training dataset.

(2) LASSO: LASSO (Tibshirani, 1996) is a linear regression model that performs L1 regularization. It has been previously applied for road traffic prediction (Kamarianakis et al., 2012).

(3) SVR-linear: Support Vector Regression (Cortes and Vapnik, 1995) models have been applied widely for short-term traffic flow prediction (L. Lin et al., 2013a; Lin et al., 2014a), travel time prediction (Wu et al., 2004), traveler socio-demographic role prediction (Zhu et al., 2017), etc. SVR with a linear kernel (SVR-linear) can greatly reduce the training time but may sacrifice prediction performance.

(4) SVR-RBF: SVR with Radial Basis Function kernel is a popular type of support vector regression models with proven capability in prediction tasks.

(5) MLP: The Multi-layer Perceptron is a classical feedforward neural network model. Note that in this study a MLP model is built for each station by revising the GCNN$_{reg}$-DDGF model, which requires little effort.

(6) LSTM: The LSTM model (Hochreiter and Schmidhuber, 1997) has proven to be stable and powerful for modeling long-range dependencies in temporal sequences. However, it is not designed to capture spatial dependencies (Ke et al., 2017). The LSTM model in this study follows the same architecture as the recurrent block in the GCNN$_{rec}$-DDGF model.

(7) XGBoost: XGBoost is a gradient boosted regression tree technique (Chen and Guestrin, 2016). In the transportation domain, XGBoost was reported as one of the most powerful algorithms in the 2014 Kaggle Bike Sharing Demand Prediction competition (Kaggle, 2015). In 2017, the winning team in the KDD Cup Competition Highway Travel Time and Traffic Volume Prediction competition used XGBoost and neural network models as the final solution (KDD Cup, 2017). The hyper-parameters of the XGBoost model are determined based on the Bayesian optimization method (Brochu et al., 2010).

The evaluation criteria for the testing dataset include the RMSE (Equation (16)), and additionally, the Mean Absolute Error (MAE) and the coefficient of determination ($R^2$) calculated as follows:

$$MAE = \frac{1}{M*N}\sum_i^M \sum_j^N |y_{ij} - P_{ij}| \tag{17}$$

$$R^2 = 1 - \frac{\sum_i^M \sum_j^N (y_{ij} - P_{ij})^2}{\sum_i^M \sum_j^N (y_{ij} - \bar{y})^2} \tag{18}$$

where $\bar{y}$ is the average demand in the testing dataset.

**Table 3**

Model Comparison on Testing Dataset

| Model | RMSE | RMSE (7:00 AM – 9:00 PM) | MAE | $R^2$ |
|---|---|---|---|---|
| GCNN$_{rec}$-DDGF | 2.12 | 2.58 | 1.26 | 0.75 |
| GCNN$_{reg}$-DDGF | 2.35 | 2.85 | 1.43 | 0.70 |
| XGBoost | 2.43 | 2.95 | 1.44 | 0.68 |
| LSTM | 2.46 | 3.00 | 1.44 | 0.67 |
| GCNN-DC | 2.50 | 3.02 | 1.53 | 0.66 |
| MLP | 2.51 | 3.05 | 1.51 | 0.65 |
| GCNN-DE | 2.67 | 3.21 | 1.60 | 0.61 |
| SVR-RBF | 2.67 | 3.25 | 1.57 | 0.61 |
| LASSO | 2.70 | 3.27 | 1.65 | 0.60 |
| SVR-linear | 2.72 | 3.31 | 1.52 | 0.59 |
| GCNN-SD | 2.77 | 3.31 | 1.68 | 0.58 |
| HA | 3.44 | 3.42 | 2.08 | 0.35 |
| GCNN-ATD | 3.44 | 3.83 | 2.21 | 0.35 |

Table 3 compares the model performances on the testing dataset. In addition to the RMSE, MAE and $R^2$ for the whole testing dataset, we calculate the RMSE for the period 7:00 AM to

9:00 PM, which is meaningful because bike-sharing demand for the other time periods are mostly zero or close to zero. As can be seen, GCNN$_{rec}$-DDGF performs the best on all criteria. It has the lowest RMSE (2.12), RMSE (7:00 AM – 9:00 PM) (2.58) and MAE (1.26), and the highest $R^2$ (0.75). GCNN$_{reg}$-DDGF performs the second best. This indicates that the design of the DDGF can capture the hidden correlations between stations to improve the prediction performance, and the importing of the recurrent block from the LSTM architecture enhances the GCNN-DDGF model by utilizing the temporal dependencies that exist in the bike-sharing demand series. The two GCNN-DDGF models are followed by XGBoost and LSTM in terms of performance. While XGBoost is not designed to capture temporal dependencies in the bike-sharing demand series or the hidden correlations between stations, it supports fine tuning and addition of regularization parameters to control overfitting, leading to its strong performance (Chen and Guestrin, 2016). LSTM performs close to XGBoost by utilizing temporal dependencies in the bike-sharing demand series. The next best performance is by GCNN-DC, in which the pre-definition of the adjacency matrix with the Pearson Correlation Coefficient based on the hourly bike demand series between stations makes it the best among the four GCNNs with pre-defined adjacency matrices. The MLP's performance being the third best among the seven benchmark models could be due to the building of a separate model for each station based on the architecture of GCNN$_{reg}$-DDGF model. The GCNN-ATD model performs the poorest, and has the largest RMSE (3.44), RMSE (7:00 AM – 9:00 PM) (3.83) and MAE (2.21), and the lowest $R^2$ (0.35). This indicates that ATD is not appropriate as a graph adjacency matrix. It also shows that the quality of the adjacency matrix has a huge impact on the performance of the GCNN model. The remaining benchmark models perform poorly as they do not factor correlations between stations or temporal dependencies in the series data. Especially, the HA model performs almost as poorly as the GCNN-ATD model with a RMSE of 3.44 and a $R^2$ of 0.35. The next section discusses a more detailed graph network analysis based on the learned DDGF to uncover the "black box" of the GCNN-DDGF model.

*5.3. Graph Network Analysis based on DDGF*

A bike-sharing network can be built by taking the learned DDGF $\hat{A}$ from the GCNN$_{reg}$-DDGF as the adjacency matrix. This graph network is visualized and analyzed using the popular tool Gephi (Bastian et al., 2009). Fig. 5(a) is the network visualization under the Geolocation layout in Gephi whereby the positions of the bike stations are determined by their latitude and longitude coordinates. Fig. 5(b) is the visualization with the Force Altas layout which aims to avoid vertex overlapping. It is worth noting the following points for a better understanding of the visualizations. For the edges, the DDGF $\hat{A}$ is normalized such that all entries fall within [0, 1]. Only 1,565 edges with weights not less than 0.15 are kept, and the edge thickness is proportional to its weight. The vertex size is proportional to the Weighted Degree (WD), which is the weighted sum of its edges including the self-connected one. Each vertex is labeled with its station name. Furthermore, the visualizations in Fig. 5. denote eight large communities with different colors. These communities are generated using the modularity optimization algorithm (Blondel et al., 2008; Lin et al., 2014b; Zhou, 2015). Modularity measures the strength of division of a network into smaller communities. It is maximized when the optimal division is found, which means that within the same community, the vertices are densely connected with each other, but they are sparsely connected with the vertices in the other communities in the network. Hence, the division of communities can help us to understand which stations identified by the learned DDGF can utilize the correlations among each other to improve the bike-sharing demand prediction. The sizes of these communities are also labeled in Fig. 5, e.g., "Community 1-55" implies that there are 55 stations in Community 1. In total, 251 out of the 272 stations are in these communities.

Fig. 5(a) shows that the communities have various shapes spatially. While the stations in Community 1 are mainly scattered in the middle part of Manhattan, some stations within the same community may be scattered far from each other. For example, station "Central Park S & 6 Ave" in the north and station "West St & Chamber St" in the south are both in Community 2. However, the spatial distance between the two stations is 6.1 miles. The stations in Community 7 are located along Lafayette Street, which is a major north-south street in New York City's Lower Manhattan. Fig. 5(b) shows that the edge weight strength is generally stronger within the same community, e.g., Communities 2 and 4. The vertices with the two largest WDs are station "Perishing Square North" and station "Grand Army Plaza & Central Park S". As can be seen, these two stations are connected with quite a few other stations.

*5.3.1. Weighted Degree Analysis*

WD is one of the most important measurements in graph analysis. It is interesting to determine which factors impact the WD in the bike-sharing graph from the DDGF. Fig. 6(a) explores the correlations between WD and total demand per station in the training dataset. The linear regression model is applied to fit the points. It shows that in general WD is larger when the total demand of the station is higher. The R squared value of the linear regression model is 0.51. Station "Pershing Square North" has the highest total demand of over 300,000 and the largest WD (8.46). Similarly, for each vertex in the bike-sharing graph from the DDGF, the number of neighbors connected to it can also be calculated. Fig. 6(b) plots the points based on WD and the neighbor number. It shows that a higher WD implies that this vertex has more neighbors, and the linear regression model fits the data points very well with a R squared value of 0.96. Station "Pershing Square North" has the largest WD and 39 neighbors. A previous study indicates that Station "Pershing Square North" is the most popular station in Citi BSS because of its convenient access to NYC's major transit hub (Grand Central Terminal) for commuters (Warerkar, 2017). The largest WD and high neighbor number indicate that the DDGF seeks to gather information from more stations to improve the hourly demand prediction for it.

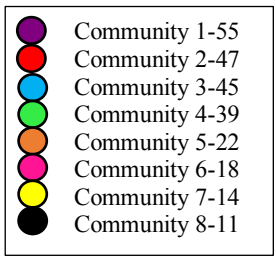
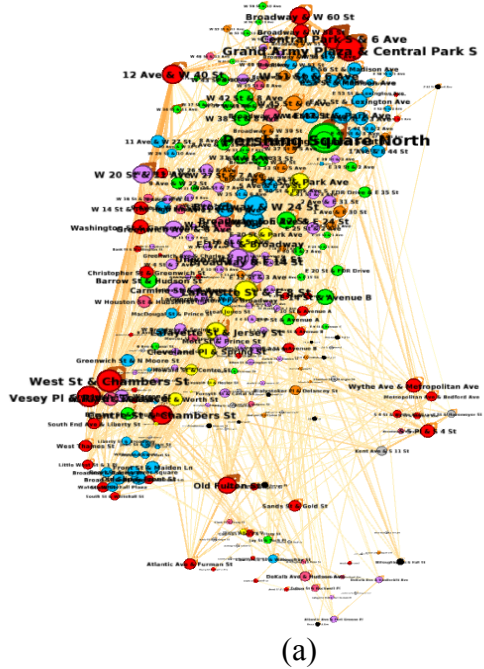

(a)

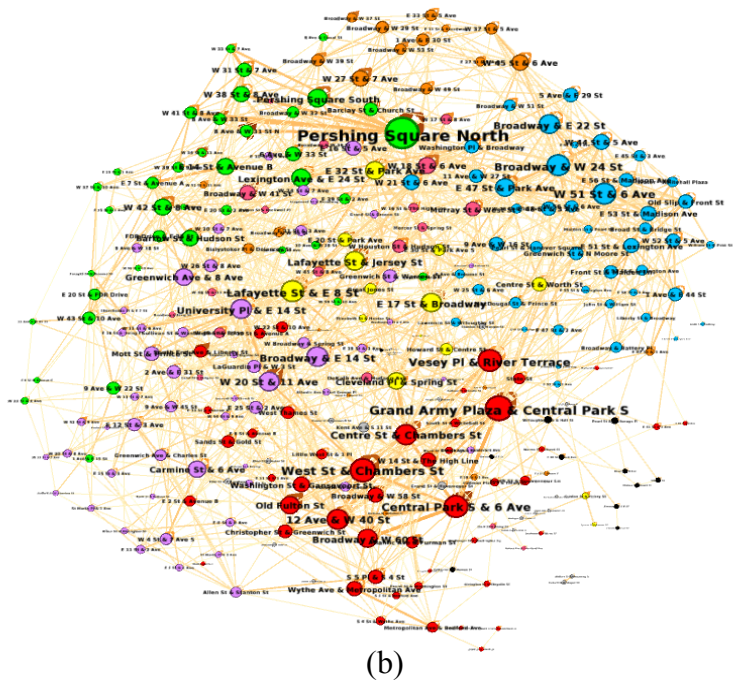

(b)

**Fig. 5.** (a) Visualization of the Bike-Sharing Network based on DDGF - Geolocation Layout, and (b) Visualization of the Bike-Sharing Network based on DDGF - Force Altas Layout.

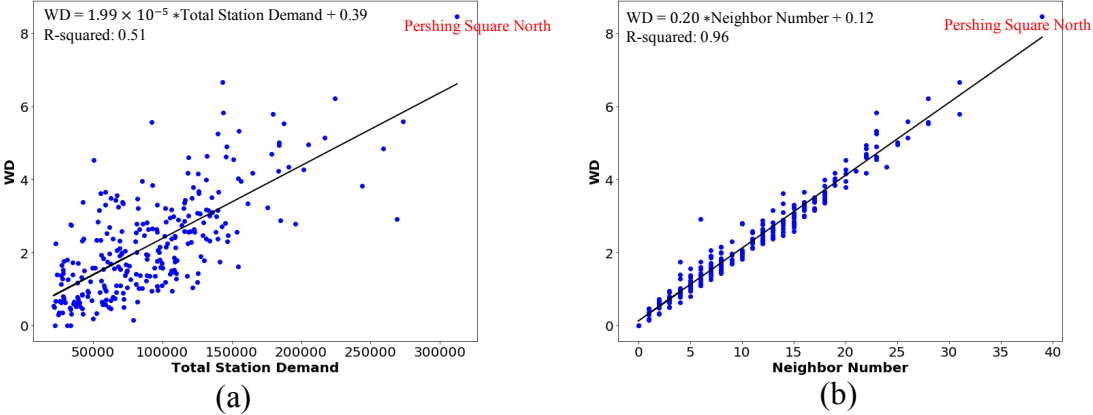

**Fig. 6.** (a) Linear Regression between WD and Total Station Demand, and (b) Linear Regression between WD and Neighbor Number.

Fig. 7(a) applies linear regression with station RMSE as the dependent variable and WD as the explanatory variable. It shows that, in general, station RMSE increases with an increase in the WD. Recall that Fig. 6 shows a small WD, implying a low total station demand; hence, it is expected that the station RMSE will also be low, and vice versa. Two stations "Pershing Square North" and "Duffield St & Willoughby St" in Fig. 7(a) are selected. The former has the largest WD (8.46) and the highest RMSE (7.39), while the latter has a WD close to zero and a very low RMSE (0.97). Fig. 7(b) and Fig. 7(c) compare one week (06/19/2016 – 06/25/2016) of predictions with the actual values for these two stations, respectively.

As can be seen, Station "Pershing Square North" is much busier, with its highest hourly demand close to 120, while it is only 10 for station "Duffield St & Willoughby St". Although the overall station RMSE of the former is much higher than that for the latter, Fig. 7(b) and Fig.

7(c) show that the predictions of the former are more accurate except for a few peak hour periods such as the afternoons on Tuesday, Wednesday, and Saturday. The overall $R^2$ values for the two stations are then calculated, which is as high as 0.72 for the former but only 0.21 for the latter. Similar to Fig. 6, this verifies that the GCNN-DDGF model can capture hidden correlations between stations to improve its prediction performance for stations with large WD. The next section will further compare the learned DDGF with other pre-defined matrices (SD, DE, and DC) to understand these hidden correlations.

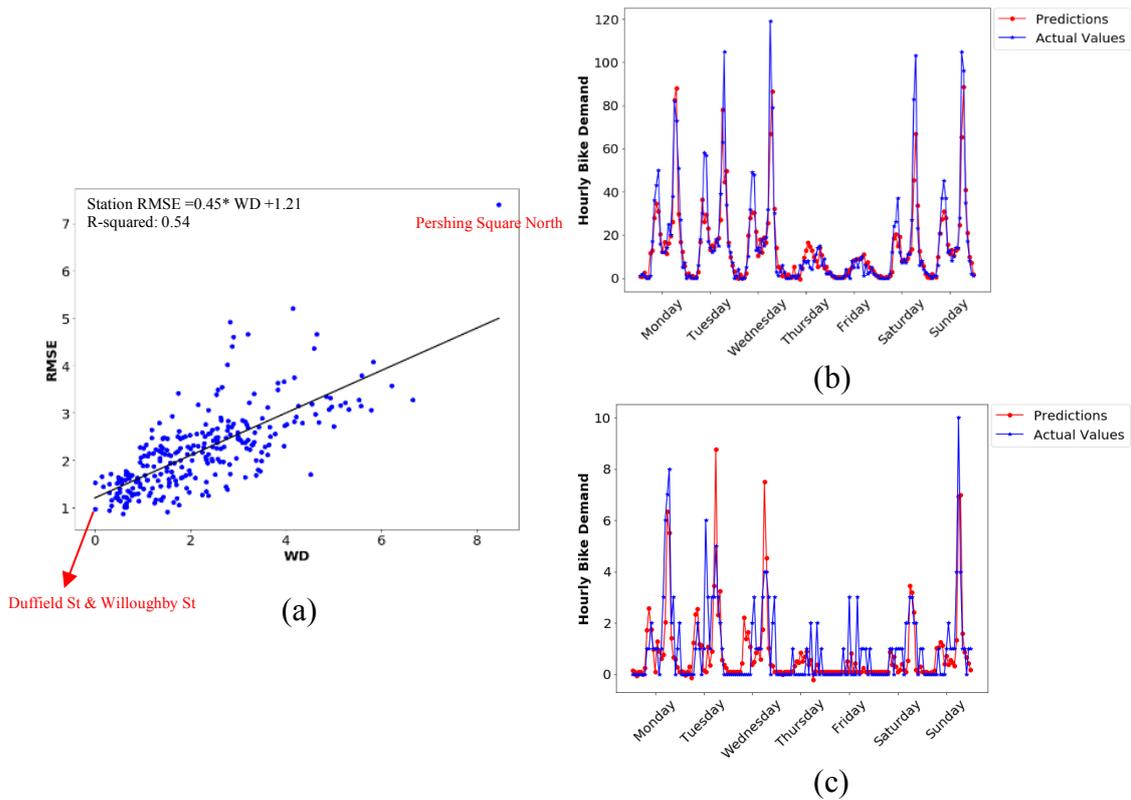

**Fig. 7.** (a) Linear Regression between Station RMSE and WD, (b) Predictions and Actual Values for "Pershing Square North" (06/19/2016 – 06/25/2016), and (c) Predictions and Actual Values for "Duffield St & Willoughby St" (06/19/2016 – 06/25/2016).

*5.3.2. Comparison between DDGF and Other Matrices*

We compare the learned DDGF with the other matrices (SD, DE, and DC) to determine why GCNN-DDGF models perform better than the other GCNNs. The ATD matrix is not considered because GCNN-ATD has very poor performance on the testing dataset.

In Fig. 8, for each of the eight communities in the bike-sharing graph from DDGF, the average of the edge weights is calculated using the distance, the demand, and the demand correlation coefficient between stations separately. Fig. 8(a) verifies that the spatial shapes of the communities are different, as shown in Fig. 5(a); e.g., the farthest distance between stations in Community 7 is only 2-3 miles, but for Communities 2 and 6, the longest one is 5-6 miles. For all communities, the average edge weight is the largest when stations are spatially close to each other (0-1 miles); after that, the average edge weight curves have a large drop when the distance changes to 1-2 miles. Beyond that, the curves are mostly steady for Communities 1, 3 and 6. However, there also exist some fluctuations for a few communities. For example, for Communities 2 and 4, the average edge weight becomes relatively higher again when it is 3-4 miles. To some extent, the DDGF is like the SD; the demand at a station has connections mainly with its spatially close neighbors; e.g., the bike demands are usually similar for a few stations near the subway in the morning peak period. However, the DDGF also reveals that the edge weight could still be large when two stations are far from each other. Therefore, the DDGF covers more heterogeneous pairwise information than the SD matrix.

Similarly, Fig. 8(b) shows that when the demand between stations is fewer than 1000, the average edge weight is the smallest for all communities, and the curves for most communities have a decreasing trend in general. Note the average edge weight curves of Communities 4 and 7 increase first and reach the maximum when the demand between stations is between 2000-3000. Beyond that range, the two curves drop with the decrease in demand. It is also worth noting that Community 2 has a few observations with demand much higher than 5000, which are excluded here to enable focus on the demand up to 5000. Its average edge weight curve has a decreasing but fluctuating trend as shown in Fig. 8(b). Fig. 8(c) illustrates that the average edge weight is the highest when the demand correlation coefficient is in the range of [0.8, 1] for all eight communities. For other demand correlation ranges, the average edge weights are much lower, and the curves are almost flat. For Community 8, the average edge weight is 0 when the demand correlation is in [0.6, 0.8) because no observations are found. Fig. 8(b) and Fig. 8(c) also illustrate that the correlations between stations based on the DDGF are consistent with the DE and the DC matrices to some extent, like the smallest average edge weight for the lowest demand range in Fig. 8(b) and the highest average edge weight for the highest demand correlation range in Fig. 8(c). However, the nonlinear curves in Fig. 8(b) and Fig. 8(c) also indicate that the DDGF covers more heterogeneous pairwise information than these matrices.

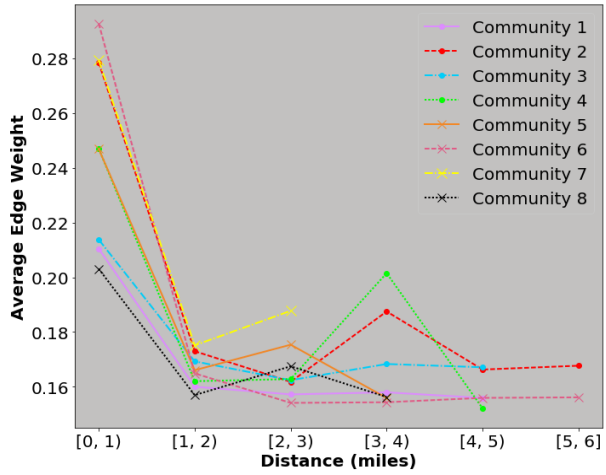

(a)

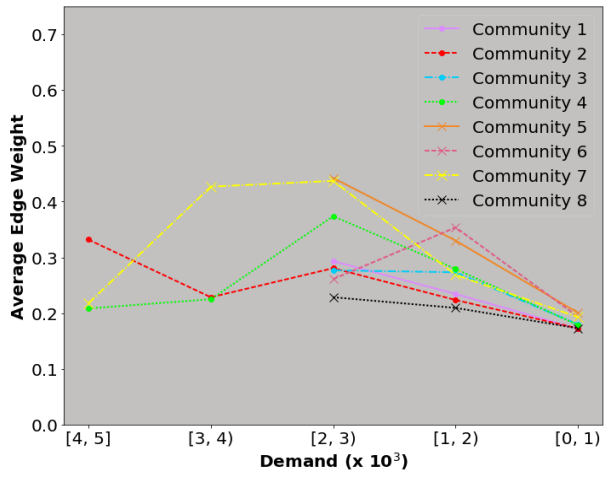

(b)

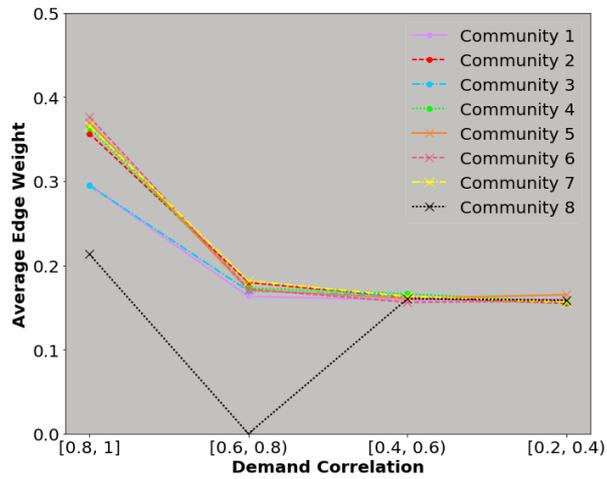

(c)

**Fig. 8.** (a) Average Edge Weight by Spatial Distance, (b) Average Edge Weight by Demand, and (c) Average Edge Weight by Demand Correlation.

To futher reveal the learned DDGF $\hat{A}$ can capture some information from the DE and DC matrices, we select stations with the four largest WDs: "Pershing Square North", "Grand Army Plaza & Central Park S", "West St & Chambers St", and "Central Park S & 6 Ave". For each station, the corresponding row in $\hat{A}$ is sorted from the largest to the lowest to rank its neighbors including itself. As shown in the subplots of Fig. 9, the first column indicates those neighbors that have the 10 largest edge weights, and their ranks. The ranks of these neighbors based on the DE and DC matrices are shown in the second and third columns. For convenience, a station square is colored green if its rank is in the top 10 list; otherwise, the color is set as red.

Fig. 9 shows that the self-connection for each station has the largest edge weight, which suggests that its own demand series plays the most important role in predicting the next hour's demand. Another observation may explain why station "Central Park S & 6 Ave" in the north and station "West St & Chamber St" in the south are strongly connected in Fig. 5. As shown in Fig. 9(d), for station "Central Park S & 6 Ave", "West St & Chamber St" is ranked $6^{th}$ based on the DDGF, $8^{th}$ based on the DC matrix, and $2^{nd}$ based on the DE matrix. This indicates that the demand series of "West St & Chamber St" can definitely help to predict the next hour's demand of station "Central Park S & 6 Ave". For station "West St & Chamber St" in Fig. 9(c), the edge weight between it and "Central Park S & 6 Ave" is ranked $3^{rd}$ among its neighbors, but the ranks based on DE and DC are only $24^{th}$ and $16^{th}$, respectively. This may indicate that the relationship between the two stations is not simply symmetric. As a future research direction, GCNN-DDGF models should be extended to the directed graph so that the edge weights between stations can cover distinct connections.

Finally, in most instances in Fig. 9 when the squares are green, the ranks based on the learned DDGF, DE, and DC are consistent. For example, for station "Central Park S & 6 Ave" in Fig. 9(d), of the neighboring stations that rank in the top 10 based on the DDGF, 7 of them are also in the top 10 lists based on DE and DC. However, sometimes the ranking based on the DDGF may not agree with that from either the DE or DC matrix. For example, for station "Grand Army Plaza & Central Park S" in Fig. 9(b), the neighbors from ranks 5 to 10 based on the DDGF have much lower ranks under the DE matrix, and 3 of them are not even in the top 10 list based on the DC matrix. Like the conclusion from Fig. 8, this suggests that the DDGF does include some information from the DE and DC matrices though none of them are used as inputs in the GCNN-DDGF algorithm. At the same time, the DDGF also encodes some hidden connections between stations that cannot be explained by the DE and DC matrices.

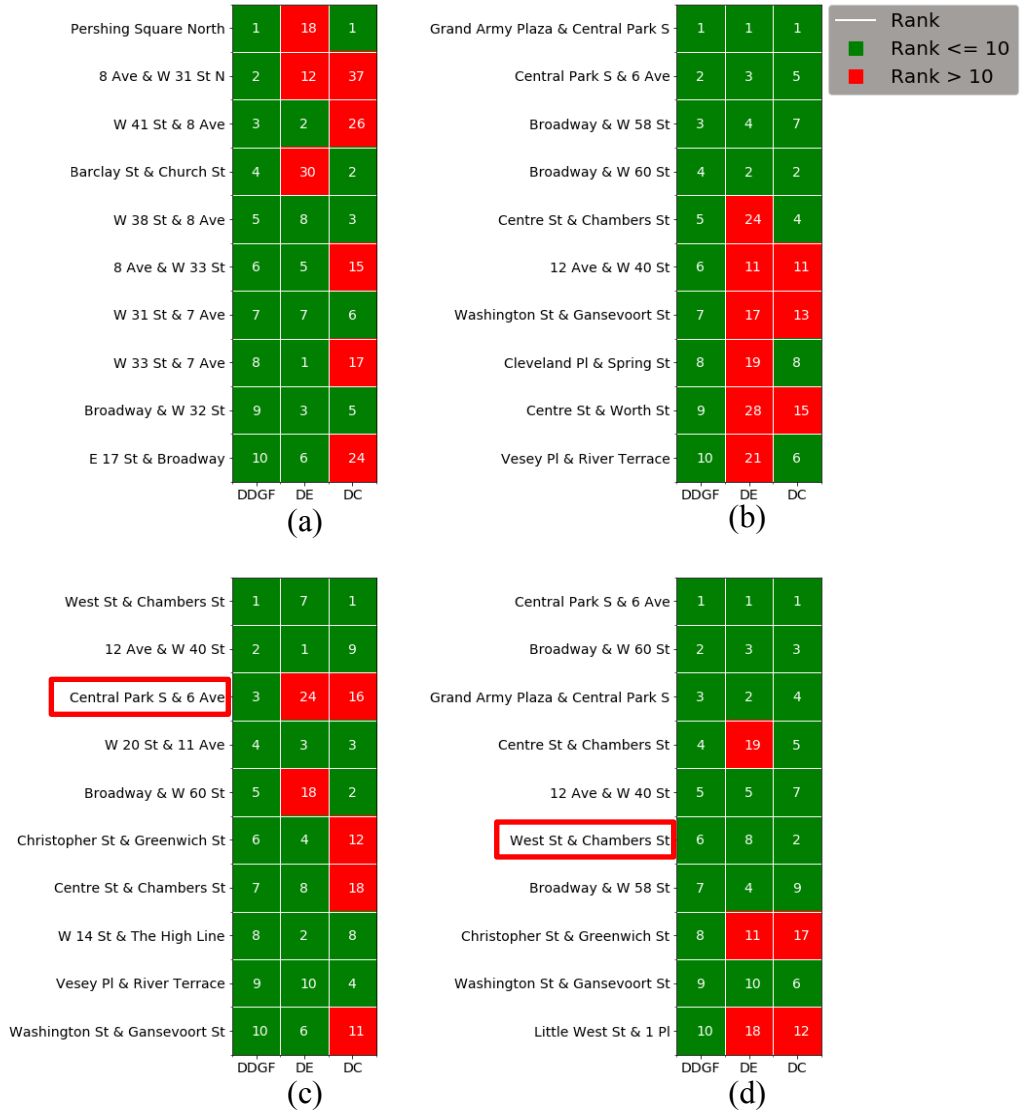

**Fig. 9.** Ranks of Neighbors for Stations with the Four Highest WDs: (a) Pershing Square North, (b) Grand Army Plaza & Central Park S, (c) West St & Chambers St, and (d) Central Park S & 6 Ave.

**6. Conclusions and Future Research Directions**

This paper proposed a novel GCNN-DDGF model for station-level hourly demand prediction in a large-scale bike-sharing network. We explored two architectures of the GCNN-DDGF model: GCNN$_{reg}$-DDGF and GCNN$_{rec}$-DDGF. Both can address the main limitation of a GCNN, that its performance relies on a pre-defined graph structure. Beyond automatically capturing heterogeneous pairwise correlations between stations to improve prediction, GCNN$_{rec}$-DDGF also implements the recurrent block from the Long Short-term Memory (LSTM) neural network architecture to capture the temporal dependencies in the bike-sharing demand series. Due to these two features, GCNN$_{rec}$-DDGF outperforms all other tested models including four GCNN models with predefined adjacency matrices and seven benchmark models. Furthermore, the bike-sharing graph network based on the learned DDGF was analyzed in detail. A key insight is that the DDGF not only captures some of the same information existing in the SD, DE and DC matrices, but also uncovers hidden correlations among stations that are not revealed by any of these matrices.

In future research, it would be meaningful to consider more variables such as weather and social events (holidays and sports games) in bike-sharing demand prediction models (Lin et al., 2015); for example, these variables can be concatenated with the input layer in the feedforward block in GCNN-DDGF models. Further, the current model can be improved to an online model which can adjust the hyper-parameters continuously. Third, the GCNN models can also be applied to solve other transportation problems that can be represented by graphs such as subway station demand prediction, network traffic state estimation, and so on. Fourth, the GCNN model can be extended to capture uncertainties in predictions (Lin et al., 2018a). Fifth, the GCNN model can be considered as a component in a comprehensive framework for dynamic bike rebalancing. Finally, it would be useful to make the model learn a sparse graph filter, as well as be applicable to a directed graph.

**Acknowledgements**

This work was partially supported by the U.S. Department of Transportation through the NEXTRANS Center, the USDOT Region 5 University Transportation Center. The authors would like to thank the Office of the Assistant Secretary for Research and Technology of the U.S. Department of Transportation.